\documentclass[twoside,11pt]{article}

\usepackage[abbrvbib, preprint]{jmlr2e}

\usepackage[utf8]{inputenc} 
\usepackage[T1]{fontenc}    
\usepackage{lmodern}
\usepackage{hyperref}       
\usepackage{url}            
\usepackage{booktabs}       
\usepackage{amsfonts}       
\usepackage{nicefrac}       
\usepackage{microtype}      
\usepackage{amsmath}
\usepackage{multicol,multirow}
\usepackage{booktabs}
\usepackage{amssymb}
\usepackage[boxruled]{algorithm2e}
\usepackage{breqn}
\usepackage{graphicx}  
\usepackage{color}

\newcommand{\lec}{{\sc LeCaR}}

\newcommand{\alec}{{\sc OLeCaR}}

\usepackage{empheq}

\newcommand{\EXP}[1]{{\sc EXP4}}
\newcommand{\ama}[1]{{\sc (AMaDAyS)}}
\DeclareMathOperator*{\E}{\mathbb{E}}

\newcommand{\qed}{\hfill \ensuremath{\Box}}

\jmlrheading{21}{2020}{1-48}{9/20}{9/20}{farzanamab}{Farzana Beente Yusuf, Vitalii Stebliankin, Giuseppe Vietri and  Giri Narasimhan}

\ShortHeadings{Delayed Feedback and Decaying Costs MAB}{Yusuf et al.}
\firstpageno{1}

\begin{document}

\title{Cache Replacement as a MAB with Delayed Feedback and Decaying Costs}

\author{\name Farzana Beente Yusuf \email fyusu003@fiu.edu \\
       \addr Algorithms for Machine Learning, Data Analytics, and Computing Systems \ama\ \\ 
       School of Computing and Information Sciences \\ 
       Florida International University, Miami, FL 33199, USA
       \AND
      \name Vitalii Stebliankin \email vsteb002@fiu.edu \\
       \addr Algorithms for Machine Learning, Data Analytics, and Computing Systems \ama\ \\ 
       School of Computing and Information Sciences \\ 
       Florida International University, Miami, FL 33199, USA
       \AND
       \name Giuseppe Vietri \email giusevtr@gmail.com  \\
       \addr Computer Science and Engineering Department\\
       University of Minnesota, Minneapolis, MN 55455, USA
       \AND
       \name Giri Narasimhan \email giri@fiu.edu \\
       \addr Algorithms for Machine Learning, Data Analytics, and Computing Systems \ama\ \\ 
       School of Computing and Information Sciences \\ 
       Florida International University, Miami, FL 33199, USA
        }

\editor{Kevin Murphy and Bernhard Sch{\"o}lkopf}

\maketitle

\begin{abstract}
  Inspired by the cache replacement problem, we propose and solve a new variant of the well-known multi-armed bandit (MAB), thus providing a solution for improving existing state-of-the-art cache management methods. Each arm (or expert) represents a distinct cache replacement policy, which advises on the page to evict from the cache when needed. Feedback on the eviction comes in the form of a ``miss'', but at an indeterminate time after the action is taken, and the cost of the eviction is set to be inversely proportional to the response time. The feedback is ignored if it comes after a threshold value for the delay, which we set to be equal to the size of the page eviction history. Thus, for delays beyond the threshold, its cost is assumed to be zero. Consequently, we call this problem \textit{MAB with delayed feedback and decaying costs}.
  We introduce an adaptive reinforcement learning algorithm \textit{EXP4-DFDC} that provides a solution to the problem. We derive an optimal learning rate for \textit{EXP4-DFDC} that defines the balance between \textit{exploration} and \textit{exploitation} and prove theoretically that the expected regret of our algorithm is a vanishing quantity as a function of time.
  As an application, we show that \lec, a recent top-performing machine learning algorithm for cache replacement, can be enhanced with adaptive learning using our formulations. We present an improved adaptive version of  \lec, called \alec, with the learning rate set as determined by the theoretical derivation presented here to minimize regret for \textit{EXP4-DFDC}. 
  It then follows that \lec\ and \alec\ are theoretically guaranteed to have vanishing regret over time.
\end{abstract}


\section{Introduction}
Software programs can be speeded up using software or hardware caches. The \emph{Cache Replacement} problem is a fundamental problem in computer science that dictates the performance of such caches.
Classical cache replacement algorithms such as \emph{Least Recently Used} (LRU) and \emph{Least Frequently Used} (LFU) are known to perform well on average but leave a lot of room for improvement. The \emph{Adaptive Replacement Cache} (ARC) algorithm showed better performance by exploiting both frequency and recency properties of data items, and quickly became the state-of-the-art method for cache replacement after its invention \citep{mm-ARC-03, mm-ARC-04}. More recently, some algorithms have demonstrated comparable or even better performance than ARC, such as LIRS \citep{jiang2002lirs} and DLIRS \citep{li2018dlirs}.
Among the recent crop of outstanding cache replacement algorithms is the \lec\ algorithm invented by \citet{vietriLecar-2018}, which uses online reinforcement learning, thus opening a promising new direction to improve cache replacement with the help of machine learning.
In this study, we aim to address the cache replacement problem by formulating a new variant of the \emph{Multi-Armed Bandit} (MAB) problem and providing theoretical insight into how to improve the \lec\ algorithm for cache replacement.
The original MAB \citep{robbins1985some} problem can be formulated as a $T$-round game where a player has access to a panel of $N$ experts, each following a specific strategy. 
In round $t$, the player consults the $N$ experts (arms), each of whom recommends one of $K$ possible actions, represented as a $K$-dimensional binary vector, $\xi_i(t)$ with only one occurrence of a 1. 
Thus $\xi_i(t) \in \{0,1\}^K$ with only one 1. The action in round $t$ is $j$ if and only if $\xi_i^j(t) = 1$, where $\xi_i^j(t)$ is the $j$-th component of $\xi_i(t)$. 
We also note that the proof can be generalized to the case when the recommendations from each expert is  probabilistic, i.e., $\xi_i^j(t) \le 1$ and the components of vector $\xi_i(t)$ add up to 1.
Each action is associated with a reward or cost value, which is provided in the form of feedback to the player immediately following the action.
The problem is to find the strategy that results in the highest gain.
A successful algorithm for MAB must strike a balance between acquiring new knowledge about the input (i.e., ``exploration'') and optimizing decisions based on existing knowledge (i.e., ``exploitation'').
Cache replacement can be considered as a variant of the MAB where, for each page request $\sigma(t)$, expert $i$ recommends an action $\xi_i(t)$ indicating the page in the cache to be evicted, if a replacement is needed. 
A feedback for the action in round $t$ is provided when the  decision evicts a page $X$, which triggers a ``miss'' when $X$ is requested at some later round. 
Therefore, the feedback is delayed by an indeterminate amount, and the magnitude of the associated cost is inversely proportional to the time passed since the action was taken. Note that because of the application being considered, it is convenient for us to use the term ``cost'' instead of ``reward'', although they are equivalent.

Even though variants of MAB with delayed feedback have been studied previously \citep{weinberger2002delayed, agarwal2011distributed, langford2009slow, neu2010online, desautels2014parallelizing, dudik2011efficient}, they cannot be applied to the cache replacement problem. 
First, in all existing studies, cost increases with increased delay. 
In our case, with larger delay, the cost associated with the chosen action decreases. 
Second, the cost vanishes when the delay is larger than a specified threshold. 
%
In this paper, we formulate a new MAB variant - \emph{MAB with delayed feedback and thresholded decaying cost} (MAB-DFDC). 

To solve the newly formulated problem, we introduce the EXP4-DFDC algorithm and show that it exhibits vanishing regret over an increasing time horizon. 
Finally, we present \alec \ -- an enhanced \lec \ version with a theoretically optimal adaptive learning rate derived from regret minimization applied to EXP4-DFDC.

\section{Related Work}
\subsection{Variants of MAB}
The classical \textit{Multi-Armed Bandit} problem was initially formulated by \citet{robbins1952some, robbins1985some}. The motivating application was to find a balance between two competing goals of finding the best treatment (exploration) and treating the patients as soon as possible from the best known treatments (exploitation) so far in clinical trails for treatments. 
The formulation of MAB in terms of a $T$-round repeated game is as follows: For each play $t$ over $T$ rounds, the player selects one of the $K$ actions, $\xi_i(t) \in \{0,1\}^K, i = 1, \ldots, n,$, where $\sum_{j=1}^K \xi_i^j(t)=1$ and obtains a cost of $x_j(t)$ for this play.
The cost for each action is associated with an unknown probability distribution.
The goal is to find the the best strategy to follow  by selecting actions that will minimize the total cumulative cost, thus minimize regret.
As a result, many different algorithms have been proposed depending on the regret minimization technique, which involves playing a mix of exploration and exploitation strategies.

\citet{lai1985asymptotically} proved that the player’s regret over $T$ rounds can be made as small as $O(\log T)$ in a stochastic setting based on the assumption that the cost distribution over the actions are random.
As $T \rightarrow \infty$, the average regret vanishes.
\citet{littlestone1994weighted} proposed a \textit{weighted majority algorithm} with $N$ experts which makes at most $c(\log |N| + m)$ mistakes on that play sequence, where $c$ is a fixed constant and $m$ is the number of mistakes made by the best expert.
The player chooses the action based on the majority voting, and the weights are updated based on the associated cost. \citet{vovk1995game} provided similar bounds on the cumulative loss of the learning algorithm for the prediction problem, where it never exceeds $c L+ a \log N$ such that $a$ and $c$ are fixed 
constants, $N$ is the number of experts, and $L$ is the cumulative loss 
incurred by the best expert in the pool. Variants of algorithms to solve a worst-case stochastic MAB problem in an online framework were developed \citep{freund1995desicion,
freund1999adaptive} based on the previous mentioned works.

In certain applications, the cost function cannot be modeled by a stationary distribution. As an example, in a communication network, to find the best route for transmitting packets from a fixed number of possible options requires sophisticated statistical assumptions on the associated costs of the routes. In other applications, it might be impossible to determine the appropriate distribution. Therefore, a \textit{non-stochastic multi-armed bandit} (or a \textit{adversarial bandit}) makes no assumption about the nature of the cost generation process \citep{Auer_2002}.
Instead of a well-defined stochastic distribution, the adversary takes control over the costs generation process. 
The adversary may be \textit{oblivious} (where the costs of all actions at all rounds are selected in advance) or \textit{non-oblivious} (where the adversary changes the distribution based on the player strategy) \citep{auer1995gambling}. 

The oblivious 
\textit{adversarial} MAB problem can be solved using the \emph{Exponential Explore-Exploit} (EXP3) algorithm \citep{Auer_2002},  which is a variant of the ``Hedge'' algorithm proposed by \citet{freund1995desicion}. ``Exponential weight algorithm for Exploration and Exploitation using Expert advice'' (EXP4) \citep{Auer_2002} extends the EXP3 algorithm for multiple experts settings.
The standard weighted majority algorithm is not effective in non-stochastic settings because experts with large weights could prevent actions with potentially smaller costs that are delivered later in time. 
Both the EXP3 and EXP4 algorithm has an exploration parameter, which controls the probability of choosing the arm at random. 
It have been shown that both EXP3 and EXP4 player strategies guarantee a worst-case regret of $O(\sqrt{T})$, where $T$ is the number of rounds played \citep{Auer_2002}.


Many variants of the MAB problem and solution approaches have been developed inspired by different applications, i.e., online advertisement, network communication modeling, and news recommendation system based on the adversarial bandit work of \citet{Auer_2002}. In some of these cases, the feedback is not instantaneous, instead delayed due to some constraints.
For example, \citet{weinberger2002delayed} studied  \textit{MAB with delayed feedback} in the adversarial full information setting (the player receives feedback information related to all the actions instead of only the chosen action). They discussed an application of this to prefetch pages in computer memory systems and proved vanishing regret. \citet{mesterharm2005line} explored another variant of the full information setting when side information is available. For example, in a social network, to recommend a new friend to a user, it is possible to exploit the mutual friend information by the recommendation system. For such a setting, the algorithm \citep{mesterharm2007improving} exhibits increasing regret to average delay for the adversarial environment. They also studied the stochastic setting and provided a regret bound, which increases with maximum delay when feedback is delayed.
\citet{langford2009slow} proved similar bounds for a sufficiently slow learner and exploited parallelism to design an online learning strategy that best uses multi-core architecture
technology to solve the problem.
\citet{neu2010online} formulated a MAB with an oblivious adversary and showed a multiplicative regret for a fixed and known delay when there is no side information available. \citet{dudik2011efficient} and \citet{desautels2014parallelizing} presented online learning algorithms
for delayed feedback resulting in a regret bound with an additive term dependent on the fixed delay and maximum delayed for no side information and side information case, respectively.

\citeauthor{beygelzimer2011contextual} proposed a solution of contextual bandits when only partial feedback is available. The work of \citet{agarwal2011distributed} and \citet{joulani2013online} on online stochastic MAB  showed that regret increases linearly with delay respectively for full and partial information delayed feedback setting. The \emph{blinded bandit} addresses another variant when the feedback is not provided on the rounds the player switches to a different action \citep{dekel2014blinded}.

One important application is the online routing network problem. During the route switching of the streaming transmission, it takes time to compute the new transmission rate. Cooperative bandit problems in multi-player settings have been studied by \citet{awerbuch2008competitive, szorenyi2013gossip} in a stochastic and adversarial setting, where the players cooperate to find the best strategy using dynamic
random networks. \citet{cesa2016delay} explored delay and cooperation in non-stochastic bandits with an application to communication networks. \citet{cesa2016delay} proved a regret bound for adversarial delayed feedback in a cooperative multi-agent setting, 
where regret increases with delay. \citet{joulani2016delay} showed regret bounds of $\sqrt{(d +T) \ln K}$, where $d$ is the total delay in feedback
experienced over the $T$ rounds with $K$ available actions for full-information settings.

\section{Problem formulation: MAB with delayed feedback}

\begin{figure*}[!ht]
  \begin{center}
    \includegraphics[width=1\textwidth]{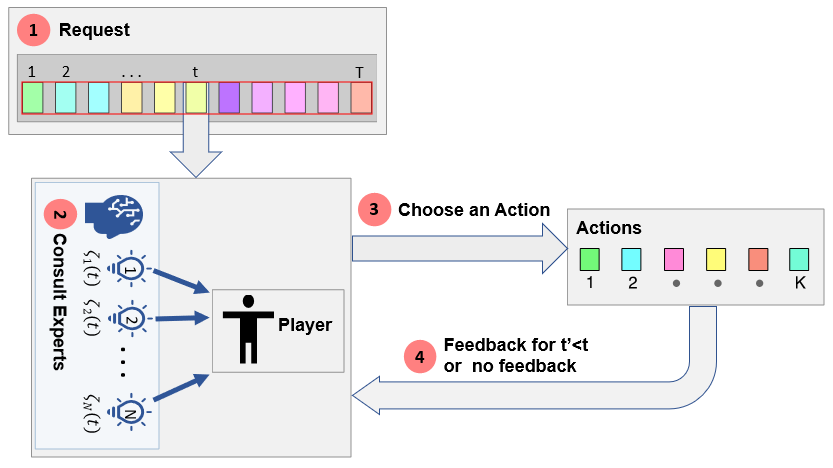}
    \caption{A schematic for the MAB-DFDC problem}
    \label{fig:1-MAB-DFDC}
  \end{center}
\end{figure*}

In this section we formulate a new \textit{MAB} variant with \textit{Delayed Feedback and Decaying Cost} (MAB-DFDC) that is applicable to the cache replacement problem.
MAB-DFDC differs from the existing literature in terms of the formulation of delayed feedback.
In the MAB problem, there is no delay in obtaining the feedback. 
In contrast, MAB-DFDC allows delays in feedback to be greater than zero. 
In previous studies that allowed delays in feedback, cost and/or regret was modeled as increasing with the delay. But, in our proposed formulation, regret decreases as delay increases. 
As inspired by the cache replacement problem, if an evicted item is requested immediately after its eviction, the regret is much higher than if the request occurs after a considerable passage of time. 
Additionally, when the delay reaches a predefined threshold, our formulation ignores its contribution and the associated cost term vanishes from the regret bound equation. The intuition here is that once the regret due to an eviction becomes appreciably small, i.e., after a sufficient passage of time, it is not worth the bookkeeping effort to track its negligible impact and can be dropped altogether from further consideration.

We describe the delayed feedback, decaying cost version of MAB problem (MAB-DFDC), and borrow from the notation used for the adversarial MAB problem by \citet{Auer_2002}.
The problem is modeled as a game where the player's objective is to minimize costs.
The player receives a sequence of requests and must respond with one of $K$ actions at the end of each request.
The player also has access to $N$ experts who are consulted on every action.
Every action has a cost and the player seeks to minimize the cost over $T$ requests as $T$ becomes large.
As described earlier, the decaying cost is unknown until the action elicits a feedback, which may be delayed for an indeterminate amount of time.
However, the cost is a pre-specified function of the delay of the feedback -- we assume a simple linear dependence.
Fig. \ref{fig:1-MAB-DFDC} provides a schematic and highlights the key concepts of the MAB-DFDC problem, from which the application to the cache replacement problem can be readily seen.
\begin{enumerate}
\item A request (e.g., a page request) is received for round $t$ in a $T$-round game.

\item The player consults with $N$ experts and picks an expert $i \in \{1, \ldots, N\}$, each of whom recommend an action $\xi_i(t)$ as a (binary) vector of probabilities, where $\xi_i(t) = (p_1^i(t), \ldots, p_K^i(t))$ and $p_j^i(t) \in [0,1], j= 1,\ldots,K$. As mentioned earlier, $\xi_i^j(t)$ can be thought of as the probability of choosing action $j$ by expert $i$ at time $t$.
Note that all experts have access to the state of the system in making their expert recommendations, and that ``no action'' is a valid recommendation. In the cache replacement problem, an expert is a page replacement policy (e.g., LRU), and the action is the associated cache manipulation. ``No action'' is recommended if the item is found in cache (i.e., a hit). Each action is associated with a cost, which is revealed to the player after an indeterminate delay.


\item An action $i_t$ is chosen by the player based on the recommendations of the experts and based on some pre-determined criteria. Therefore, $\xi_i^j(t) = 1$, if and only if $i_t = j$, and 0 otherwise. A ``history'' of actions is maintained to provide feedback in the next step. This information may also be used by the experts to provide a recommendation. For the cache replacement scenario, a \emph{hit} triggers ``no action'', but a \emph{miss} results in an action that evicts a page resident in the cache to make room for the page requested in round $t$. Eviction information is stored in history to generate a feedback in the later rounds if needed.

\item Finally, a feedback is generated. The feedback may be for the current request or for a request from some previous round $t'$, and its computation requires the information stored in history. For the cache replacement problem, a hit triggers a ``no action'' and results in an immediate feedback with no cost. if a ``miss" is encountered for the currently requested page, we search the page eviction history to find the round $t'$ when the page was last evicted from the cache.
If the currently requested page is found in history, we provide the feedback for the action from round $t'$. If the requested page is not found in history, no feedback is provided.
\end{enumerate}

The classical MAB problem assumes that $t=t'$, i.e., that the feedback is immediate for every action. 
In our case, the delay, $d = t - t'$, is defined as the number of the rounds between when the action taken and when the feedback was received for that particular action.
We assume that the cost is a function of the delay.
Consequently, its computation may require an infinite history. Since this is impractical in real applications, we assume that the history is truncated and is of bounded size.
In the cache replacement application, the history size is bounded by some memory constraint and we assume that beyond a certain delay, the real cost is negligible and can be ignored.
In other words, we do not distinguish between a request that is appearing for the first time (and therefore triggers a miss) and a request that appears after an extremely long time well after the item was last evicted from the cache.

Feedback is \emph{delayed}, and the penalty/cost for an action is a decaying function of the delay $d$. 
Also, if the delay (in feedback for any action) is higher than some predefined threshold, then it is assumed to be infinite, thus making the feedback cost for that particular action to be negligible and ignored totally.
 
Many variants of MAB have been proposed based on the way the cost function is defined.
We assume that in round $t$, the vector of values of the cost function for each action are denoted by $\mathbf{x}(t)= (x_1(t),\dots ,x_K(t))$, where $x_j(t) \in [0, 1] $. 
In an oblivious model, the cost function does not depend on the player's action in the previous rounds.
In an adversarial model, the cost function may change based on the previous actions.
In each round, no feedback is provided on the other $K-1$ actions that were not chosen by the player. 
The main objective of the player is to minimize the cost (or maximize the reward) over a $T$-round game.
The cumulative cost after $T$ rounds for the best strategy is given by 
\begin{equation*}
C_{best}(T) = \min\limits_{1 \le i \le N}\sum_{t=1}^{T} \xi_i(t) \cdot \mathbf{x}(t),
\end{equation*}
which assumes that the best of the $N$ expert recommendations are followed in each round.  
Also, the cumulative cost incurred by an algorithm $A$ selecting action $a(t)$ in round $t$ is given by
\begin{equation*}
C_{A}(T)= \sum_{t=1}^{T}x_{i_t}(t).
\end{equation*}
%
To measure the performance of adaptive learning algorithms, the concept of \emph{regret} was borrowed from the literature on the theory of games.
For any algorithm, \emph{regret} can be defined in terms of cumulative difference between the costs of the best strategy in each step and the algorithm in consideration.
Thus, the regret $ R_A(T) $ of any algorithm $A$ after round $T$ can be calculated from Eq. (\ref{eqn:regret}) below.
\begin{equation}
\label{eqn:regret}
R_A(T) =  C_{best}(T) - C_A(T).
\end{equation}
The main objective of any adaptive learning algorithm is to minimize the total regret over time, while ensuring that it vanishes over a long time horizon.

\section{Algorithms}
We first describe the EXP4 algorithm designed by \citet{Auer_2002}
before describing the EXP4-DFDC algorithm analyzed in this paper. EXP4-DFDC extends the EXP4 algorithm for the \textit{Delayed Feedback and Decaying Cost} settings.

\subsection{Existing approach: EXP4 algorithm}
The ``Exponential weight algorithm for Exploration and Exploitation using Expert advice'' (EXP4) algorithm for the non-stochastic multi-armed bandit problem assumes the existence of
a fixed set of $N$ strategies/experts.
At each round $t$, all $N$ experts are consulted.
In each round, the player either follows the advice of one of the experts or explores by picking a random action.
If no prior knowledge is available about the experts, EXP4 initializes by assigning equal weights. Thus, $w_i(1)=1, i =1,\ldots, N$.
The probability, $p_j(t)$, of picking action $j=\{1,\dots, K\}$ in round $t$ is proportional to the sum of the weights of the experts that recommend action $j$. If more experts recommend an action, that action will have higher probability to be chosen by the algorithm. This is interpreted as the ``exploitation'' of the information accumulated by the learning algorithm so far. If the player chooses to explore with a random action, the algorithm will choose an action randomly from the available options. 
In other words, $p_j(t)$, the probability of the player taking action $j$ in round $t$ is given by Eq. (\ref{eqn:action_prob}) below, for $j=\{1,\dots, K\}$.
\begin{equation}
\label{eqn:action_prob}
p_j(t) = (1-\eta) \sum_{i=1}^N \frac{  w_{i}(t) \times \xi_i^j(t)}{W_t} + \frac{\eta}{K},
\end{equation}
where $\eta$ is the learning rate,  $w_i(1) = 1, i =1,\ldots, N,$ and $W_t = \sum_{i=1}^N w_i(t)$. The learning rate $\eta$  controls the amount of exploration and exploitation at each round. If the learning rate is too high, the algorithm will explore more and exploit less and vise versa.

Whenever some feedback is made available, the algorithm updates the weights according to Eq. (\ref{eqn:weight_update}) below. 
In the EXP4 algorithm, the estimated cost vector is denoted by $\hat{\mathbf{x}}(t)$, where each element $\hat{x}_j(t)$ of the vector  is set to 
$x_j(t)/p_j(t)$ for $j=1 \ldots K$ and upper bounded by the maximum possible value of actual cost $x_j(t)$ as in \citet{Auer_2002}.  
The estimated cost $\hat{x}_{i_t}$ of any action $i_t$ ensures that the actions with low probabilities are adjusted when they get picked in a later round. Otherwise, the algorithm will keep ignoring an action because the cost was high in an earlier round. In a later round, when the cost associated with that action becomes lower, it is not reflected, and the action never gets chosen due to its low probability. 
Also, given a sequence of random choices $i_1, i_2, \ldots, i_{t-1}$ of the previous rounds, the expectations of estimated cost of any random action $i_t$ taken at round $t$ is guaranteed to be equal to the actual cost of that action. Consequently,  $\E\left[ \hat{x}_{i_t}(t) | i_1, \ldots, i_{t-1}\right] = x_{i_t}(t)$
As in \citet{Auer_2002}, the weight update rule is as follows:
\begin{equation}
\label{eqn:weight_update}
w_i(t+1) = w_i(t) \text{ exp } \left(-\frac{\eta \hat{\mathbf{x}}(t) \cdot \xi_i(t)}{K}\right), \qquad i=1,\dots, N.  
\end{equation}
The same steps are  applied repeatedly for $T$ steps, allowing us to bound the regret for \EXP{4}\ as outlined in \citet{cesa2006prediction,Auer_2002}.
\begin{theorem} \citep{Auer_2002} For any $K,T > 0$, for any learning
rate $\eta \in (0,1]$, for any family of N experts (including an
uniform expert), and for any assignments of arbitrary costs, the following holds, 
\begin{equation}
\begin{split}
R_{EXP4}(T)  \le (e -1) \eta T + \frac{K \ln N}{\eta}.
\end{split}
\end{equation}
\label{thm:auer}
\end{theorem}
Classical EXP4 cannot be used in the bounded and delayed feedback  settings because the update rule cannot be applied to rounds where no feedback is available or the feedback is delayed beyond a defined limit.
\citet{dudik2011efficient} presented an algorithm devising a modification of EXP4 for \textit{MAB with stochastic delayed feedback}.
Meanwhile \citet{neu2010online} showed a multiplicative regret for the adversarial bandit case without any additional information.
The full information case was developed by
\citet{joulani2016delay}, showing regret bounds of $\sqrt{(d +T) \ln K}$, where $d$ is the total delay in feedback experienced over the $T$ rounds.
Also the online learning case under delayed feedback for stochastic bandit optimization was analyzed by \citet{desautels2014parallelizing} and resulted in a regret bound involving a multiplicative increase that is independent of the delay and an additive term depending on the maximum delay. 

\subsection{New approach: EXP4-DFDC Algorithm}
\label{subsec:mab-dfdc}

We describe, EXP4-DFDC, a variant of EXP4, to solve MAB-DFDC, the MAB problem with delayed feedback and decaying cost. 
%
%
%
In  EXP4-DFDC, when feedback is non-zero, the weights of the experts get updated. 
A feedback is generated at round $t$, triggered by an action, $i_{t'}$, taken back in round $t'$. 
As the delay in feedback, $t-t'$, increases, the relative cost of the chosen action decays. 
Since regret is proportional to cost, the weights of experts who chose the action $i_{t'}$ are decreased using the estimated cost value observed at round $t$. Estimated cost is calculated as: 
\begin{equation}
\hat{\mathbf{x}}_{i_t}(t)= \frac{{x_{i_t}(t)}}{ d \times p_{i_t}(t)}.
\label{eqn:estimated_reward}
\end{equation}
Hence, the estimated cost calculation, which incorporates the decaying cost with delayed feedback  in EXP4-DFDC is the major difference from the existing literature in the proposed algorithm. 
Finally, we want to draw a conclusion on the performance of the generic EXP4-DFDC and specific \alec\ algorithm. 
Below we will prove a theorem for the delayed feedback and decaying case along the lines of Theorem \ref{thm:auer}.

\begin{theorem}  For any $K,T > 0$, for any learning rate, $\eta \in (0,1]$, for any family of $N$ experts, and for any assignment of arbitrary costs decaying with delay $d$, the expected regret of the algorithm can be upper bounded by: 
$
R_{\text{EXP4-DFDC}}(T) \le  2\eta T + \frac{K\ln N}{\eta}.
$
\label{thm:exp4dfdc}
\end{theorem}
\begin{algorithm}[!hbt]
\SetAlgoLined
\KwIn{Learning rate $\eta \in (0,1]$, delay $d$, delay threshold for feedback $m$ }
    \Begin{
      Set $w_i(1) = 1, \text{for }  i =1,\ldots, N$\;
      
      \For{$t = 1, 2,\dots $}{
         Obtain expert advice  $ \xi_1(t),\ldots, \xi_N(t)$\;
         Set $W_t= \sum_{i=1}^N w_i(t)$\;
         
         \For{$j = 1,\ldots, K$}{
			$p_j(t) = (1-\eta) \sum_{i=1}^N \frac{w_i(t) \times \xi_i^j(t)}{W_t}  + \frac{\eta}{K} $\;
         }
        
         Select action $i_t \in [1, K]$ using probability distribution $p_j(t)$\;
         
          \For{$j = 1,\dots,K$}{
          \begin{align*}
            \hat{x}_j(t) &=
            \begin{cases}
              \frac{x_j(t)}{ d \times p_j(t) }
              , & \text{if}\ 
              i_{t'} = j \text{; } 1 \le d \le m\\
              0, & \text{otherwise}
            \end{cases} 
             \end{align*}
        	
         }
          
         \For{$i = 1,\dots,N$}{
          \begin{align*}
             w_i(t+1) &= w_i(t) \text{ exp} \left(-\frac{\eta\hat{\mathbf{x}}(t) \cdot \xi_i(t)}{K} \right) \;        
             \end{align*}
        	
         }
      }
   } 
\caption{EXP4-DFDC}
\label{alg:one}
\end{algorithm}
{\bf Proof: } 
As with other reinforcement algorithms, we formalize the regret bound for EXP4-DFDC. For the sake of completeness, we provide here the complete steps following the derivation for \EXP{4}\ \cite{Auer_2002} regret bound and adapt it for our delayed and decaying cost setting. 

Note that $W_t= w_1(t) + w_2(t) + \ldots + w_N(t)$.
We consider the ratio of the sum of weights over all iterations.
\begin{equation}
\begin{split}
\frac{W_{t+1}}{W_t} &=  \sum_{i=1}^{N} \frac{ w_i(t+1)}{W_t}  
					= \sum_{i=1}^{N}  \frac{w_i(t)}{W_t} \exp \left(- \frac{\eta}{K} \hat{\mathbf{x}}(t) \cdot \xi_i(t) \right)
\end{split}
\label{eq:ratio1}
\end{equation} 
Replacing $\frac{w_i(t)}{W_t}$ by $q_i(t)$ and  setting  $ y_i(t) = -\hat{\mathbf{x}}(t) \cdot \xi_i(t) $ in Eq. (\ref{eq:ratio1}) results in the following equation:
\begin{equation}
\begin{split}
\frac{W_{t+1}}{W_t} = \sum_{i=1}^{N}  q_i(t) \exp \left( \frac{\eta}{K} y_i(t) \right).
\end{split}
\label{eq:ratio2}
\end{equation} 
Using well-known inequalities, $ e^x \le 1+x+ \frac{1}{2}x^2$ for  $x\le 0$, and $\frac{1}{2}< e-2$, we replace the $e^x$ term in Eq. (\ref{eq:ratio2}) by $ e^x \le 1+x+ (e-2)x^2$, giving us the following inequality.
\[
\begin{split}
\frac{W_{t+1}}{W_t}
& \le  \sum_{i=1}^{N} q_i(t) \left[ 1 +\frac{\eta}{K} {y_i}(t) +  (e-2)\left( \frac{\eta}{K} {y_i}(t) \right)^2\right] \\
& \le   \left[ 1 +\frac{\eta}{K} \sum_{i=1}^{N} q_i(t){y_i}(t) +  (e-2) \left( \frac{\eta}{K}\right)^2\sum_{i=1}^{N} q_i(t){y_i}^2(t) \right].
\end{split}
\] 
%
%
Setting $x=\frac{\eta}{K} \sum_{i=1}^{N} q_i(t){y_i}(t) +  (e-2) \left( \frac{\eta}{K}\right)^2\sum_{i=1}^{N} q_i(t){y_i}^2(t)$, applying the inequality, $1+x \le e^x $, and taking logarithms on both sides, we get
%
%
\begin{equation}
\begin{split}
 \ln\frac{W_{t+1}}{W_t} \le \frac{\eta}{K} \sum_{i=1}^{N}  q_i(t) {y_i}(t) +  \frac{(e-2){\eta}^2 }{K^2}  \sum_{i=1}^{N} q_i(t) {y_i}^2(t).
\end{split}
\end{equation}
Now, adding up for $t=1,2,\ldots,T$, we get
\begin{equation}
\begin{split}
\ln \frac{W_{T+1}}{W_1} \le \frac{\eta}{K} \sum_{t=1}^{T}\sum_{i=1}^{N}  q_i(t) {y_i}(t)  + 
(e-2) \left(\frac{\eta}{K}\right)^2 \sum_{t=1}^{T}\sum_{i=1}^{N} q_i(t) {y_i}^2(t).
 \label{eqn:log_less}
\end{split}
\end{equation}
Focusing on the weight update for one expert $i$ we get,
\begin{equation}
\begin{split}
w_i(T+1) = w_i(T) e^{\frac{\eta}{K} {y}_i(T)} = \prod_{t=1}^T e^{\frac{\eta}{K} {y}_i(t)} = \exp\left({\frac{\eta}{K} \sum_{t=1}^{T} {y}_i(t)}\right).
\label{eqn:log_g}
\end{split}
\end{equation}
Since $W_{T+1} \geq w_i(T+1)$ for all experts, $i$, we get
\begin{equation}
\begin{split}
\ln \frac{W_{T+1}}{W_1} &\geq  \ln \frac{w_i(T+1)}{W_1} = \frac{\eta}{K}\left(\sum_{t=1}^{T}{y_i}(t)\right)-\ln N.
\label{eqn:log_greater}
\end{split}
\end{equation}
Combining Eqs. (\ref{eqn:log_less}) and (\ref{eqn:log_greater}), we get the following inequality.
\begin{equation}
\begin{split}
 \sum_{t=1}^{T}\sum_{i=1}^{N}  q_i(t) {y_i}(t) &\geq   \sum_{t=1}^{T} {y_i}(t)  - \frac{K \ln N}{\eta}  - (e-2)\frac{\eta}{K} \sum_{t=1}^{T}\sum_{i=1}^{N} q_i(t) {y_i}^2(t).
\label{eqn:expert_inequality}
\end{split}
\end{equation}
Replacing back ${y_i}(t)$ by $-\hat{\mathbf{x}}(t).\xi_i(t)$ on the part of left hand side in Eq. (\ref{eqn:expert_inequality}), we get 
\begin{align}
\sum_{i=1}^{N}  q_i(t)  {y_i}(t) &=  - \sum_{i=1}^{N} q_i(t) \hat{\mathbf{x}}(t).\xi_i(t) = -\sum_{i=1}^{N} q_i(t) \sum_{j=1}^{K} \xi_i^j(t)\hat{x}_j(t).
\label{eqn:left_term_inequality}
\end{align}
We know that $\xi_i^j(t) = 1$ if $i_t = j$, and 0 otherwise.
%
Using the value of $\sum_{i=1}^{N}  q_{i}(t) \times \xi_i^j(t)$ from Eq. (\ref{eqn:action_prob}) and replacing it in Eq. (\ref{eqn:left_term_inequality}), we get:
\begin{align*}
\sum_{i=1}^{N}  q_i(t)  {y_i}(t) &= - \sum_{j=1}^{K} \left(\sum_{i=1}^{N}  q_{i}(t) \times \xi_i^j(t)) \right) \hat{x}_j(t) 
\\ &= - \sum_{j=1}^{K} \left(   \frac{p_j(t)-\eta/K}{1- \eta} \right) \hat{x}_j(t)
\\& = - \sum_{j=1}^{K} \left( \frac{p_j(t)\hat{ x}_j(t)}{ 1- \eta}- \frac{\eta \hat{x}_j(t)}{K( 1- \eta)} \right)  
\\& \le  \sum_{j=1}^{K} \frac{\eta\hat{x}_j(t)}{K(1-\eta)}
\\& = \frac{\eta\sum_{j=1}^{K}\hat{x}_j(t)}{K(1-\eta)} 
\end{align*}
Since only action $i_t$ is chosen at time $t$, we know that $\sum_{j=1}^{K} \hat{x}_{j(t)} = \hat{x}_{i_t}(t)$. Also, we know that $\eta \le 1$ and $K \ge 1$ resulting in the following inequality,
\begin{align*}
\sum_{i=1}^{N}  q_i(t)  {y_i}(t)   \le  \sum_{j=1}^{K} \frac{\eta\hat{x}_j(t)}{K(1-\eta)}  \le  \frac{\hat{x}_{i_t}(t)}{(1-\eta)} 
\end{align*}
Summing over  $t=1,2,\ldots,T$, we get
\begin{align}
\sum_{t=1}^{T}\sum_{i=1}^{N} q_i(t){y_i}(t) \le \sum_{t=1}^{T} \frac{\hat{x}_{i_t}(t)} { ( 1-\eta)}.
\label{eqn:expert_inequality_y_i}
\end{align}
Similarly,
\begin{equation}
\begin{split}
\sum_{i=1}^{N}  q_i(t){y_i}(t)^2  =   q_i(t) \left( - \hat{\mathbf{x}}(t) \cdot \xi_i(t) \right)^2  \le  \hat{x}_{i_t}(t)^2 \frac{p_{i_t}(t)}{(1-\eta)}
\le  \frac{\hat{x}_{i_t}(t)}{(1-\eta)}
\label{eqn:expert_inequality_y_i_2}
\end{split}
\end{equation}
Combining  inequalities \ref{eqn:expert_inequality}, \ref{eqn:expert_inequality_y_i} and \ref{eqn:expert_inequality_y_i_2}, we get 
\begin{equation}
\begin{split}
 \sum_{t=1}^{T} \frac{\hat{x}_{i_t}(t)}{1 -\eta}  &\geq \sum_{t=1}^{T} {y_i}(t) - \frac{K\ln N}{\eta}   -  (e-2)\frac{\eta}{K} \sum_{t=1}^{T} \frac{\hat{x}_{i_t}(t)} {1-\eta}
\end{split}
\end{equation}
Multiplying both sides with $(1-\eta)$ of the above Eq. and taking expectations, we obtain 
\begin{equation}
\begin{split}
\E\left[\sum_{t=1}^{T} \hat{x}_{i_t}(t)\right] & \geq  (1-\eta )\E\left[\sum_{t=1}^{T} y_i(t)\right] -  (1-\eta) \frac{K\ln N}{\eta} - (e-2)\frac{\eta}{K} \E\left[\sum_{t=1}^{T} \hat{x}_{i_t}(t)\right].
\label{eqn:expectation}
\end{split}
\end{equation}
Since $\E\left[ \sum_{t=1}^{T} \hat{x}_{i_t}(t) \right] = \sum_{t=1}^{T} x_{i_t}(t) = C_{\text{EXP4-DFDC}} \geq C_{best}$, we get the following inequality.
\begin{equation}
\begin{split}
C_{\text{EXP4-DFDC}} & \geq  (1-\eta )\E\left[\sum_{t=1}^{T} y_i(t)\right] -  (1-\eta) \frac{K\ln N}{\eta} - (e-2)\frac{\eta}{K} C_{best}.
\label{eqn:expectation_exp}
\end{split}
\end{equation}
Multiplying both sides of the Eq. (\ref{eqn:expectation_exp}) with -1,
%
and using the fact that $\E\left[\sum_{t=1}^{T}y_i(t)\right] = -\E\left[\sum_{t=1}^{T}\hat{x}_{i_t}(t)\cdot \xi_i^{i_t}(t)\right]= -\sum_{t=1}^{T}x_{i_t}(t) \leq C_{best}$, we get
\begin{equation}
\begin{split}
-C_{\text{EXP4-DFDC}} &\le (\eta -1)C_{best} + \frac{K\ln N}{\eta}
+ (e-2)\eta C_{best} \\&\le (\eta-1) C_{best} + \frac{K\ln N}{\eta}
+ \eta C_{best}.
\label{eqn:regret_cost}
\end{split}
\end{equation}
Finally, adding $C_{best}$ on the both sides of the Eq. (\ref{eqn:regret_cost}), we get
\begin{align*}
 C_{best} - C_{\text{EXP4-DFDC}}  &\le (\eta -1 +\eta +1)\eta C_{best}+  \frac{K\ln N}{\eta} 
 \le 2\eta C_{best} + \frac{K\ln N}{\eta}  .
\end{align*}
For a fixed time horizon $T$, since no action can result in a cost greater than $1$, $C_{best} \le T$, thus giving us the following bound on the regret, $R_{\text{EXP4-DFDC}}(t)$, for the EXP4-DFDC algorithm, completing the proof:
\begin{equation}
\begin{split}
\label{eqn:regret_bound}
R_{\text{EXP4-DFDC}}(T)   \le 2\eta T + \frac{K\ln N}{\eta}.
\end{split}
\end{equation} \qed 
%

\paragraph{Picking the optimal learning rate for vanishing regret:} Theorem \ref{thm:exp4dfdc} shows that the cumulative regret is a function of the learning rate. We argue that picking the right learning rate can minimize regret and ensure that it vanishes. 
To minimize the regret, 
we differentiate the above quantity and set it to 0. 
Thus the equation, $R_A'(T) = 2T - \frac{K\ln N}{\eta^2} = 0$, gives us the optimal value of $\eta$ as follows:
%
%
%
\begin{equation} 
\eta_{OPT} =  \min \left(1, \sqrt{\frac{K\ln N}{2T}}\right).
\label{eq:opteta}
\end{equation}
Finally, plugging this back into Eq. (\ref{eqn:regret_bound}), gives us the following regret bound:
\begin{equation} 
R_A(t) \le  2\sqrt{2KT\ln N}. 
\end{equation}
The significance of this regret bound is that as $T \rightarrow \infty$, the regret bound vanishes with the time horizon.


\subsection{Application to Cache Replacement - Analysis of the \lec \ algorithm}

In this section, we show how to use the EXP4-DFDC algorithm and its theoretical analysis from Section \ref{subsec:mab-dfdc} to improve a state-of-the-art cache replacement algorithm.
The recent \lec\ algorithm of \citet{vietriLecar-2018} is an outstanding cache replacement algorithm that is based on reinforcement learning and regret minimization.
The algorithm accepts a stream of requests for memory pages and decides which page to evict from a cache when a new item is to be stored in the cache following a ``cache miss''. 
\lec\ has been shown to be among the best performing cache replacement algorithms in practice \citep{vietriLecar-2018}. 
Experiments have shown that it is competitive with the best cache replacement algorithms for large cache sizes, and is significantly better than its nearest competitor for small cache sizes including the state-of-the-art methods like ARC, which was 
designed over 15 years ago by \citet{mm-ARC-03, mm-ARC-04}). 
The \lec\ algorithm is an online reinforcement learning algorithm that relies on only two very fundamental cache replacement policies typically taught in an introductory Operating Systems class, namely the \emph{Least Recently Used} (LRU) policy and the \emph{Least Frequently Used} (LFU) policy. 
\lec\ assumes that the best strategy at any given time is a probabilistic mix of the two policies and attempts to ``learn'' the optimal mix using a regret minimization strategy. 
Thus, the \lec\ algorithm can be thought of as a special case of MAB with two arms, but with delayed feedback and decaying costs. 

If an evicted page is requested again, then some ``regret'' is associated with the eviction decision that caused the miss. 
However, the regret is greatest if it is requested immediately after an eviction. 
As the gap between when the page is evicted and when it is requested again increases, the regret decays, until a certain threshold on the delay, beyond which the regret is assumed to drop to zero.
Regret is assigned to the policy that caused the eviction of that entry.
In other words, \lec\ attempts to sense which of the two arms is likely to result in less regret at any given point in the request sequence, and also successfully sensing when the tide may be changing to a different arm.
One of the primary shortcomings of \lec\ is that its learning rate had to be fixed.
\citet{vietriLecar-2018} experimented with different learning rates and picked one that worked best for the data sets they used.
Although the data showed that different (fixed) learning rates resulted in the best performance for different data sets, they identified a value that worked best for (most) of their data sets.
We show that the \lec\ algorithm, when modified appropriately, is exactly the EXP4-DFDC algorithm, allowing us to use the results of Section \ref{subsec:mab-dfdc} and proving that a modified version of \lec\ has vanishing regret over time. 
For the rest of this paper, the version of \lec\ with minor modifications will be referred to as \alec, which stands for \lec\ with \emph{optimal} learning rate.

The following notation will be used for the discussion below.
Let $N$ denote the number of experts. In \lec\ only two experts were exploited -- LFU and LRU.
Let $K$ denote the number of possible decisions to choose from. In \lec, this is bounded by the cache size (plus 1 for ``no action''), since the decision refers to which item to evict.
Since \lec\ manages a first-in-first-out history data structure to track the most recent evictions, the delay in feedback, denoted by $d$, is ``approximated'' by its position in the history data structure. If the entry is present in history, the delay is bounded by the size of the history, $h$. If it is not present, then the feedback is ignored, with the assumption that regret is negligible.
The delay value for the current request is used to update the weights of the experts in each round.
The user-defined learning rate is denoted by $\eta$.
The cost of an eviction in \lec\ was calculated as $x_{i_t}(t)^d$, where $x_{i_t}(t)$ was called the discount factor and was set to $0.005^{\frac{1}{K}}$.
To conform to the new formulation of EXP4-DFDC, we changed the estimated cost function to be $\frac{x_{i_t}(t)}{d}$ in \alec.
Both \lec\ and \alec\ attempt to minimize the cumulative cost given by the expression, $C_A(T)= \sum_{t=1}^{T}x_{i_t}(t)$. This, in turn, attempts to minimize the regret function given by the expression, $R_{A}(T)= C_{best}(T) - C_{A}(T)$.



      
         

        

       
        	
      

\begin{algorithm}[!hbt]
\KwIn{Request sequence, $\sigma$; Cost vectors, $\mathbf{x}(t)$; Learning rate, $\eta \in (0,1]$; History size, $h$; Cache size, $K$; set of experts $= \{LFU, LRU\}$ }

 \Begin{
 	  Set $N=2, w_1(1) = w_2(1) = 1$, and $W_t = \sum_{i=1}^N w_i(t)$\;
      
      \For{$t = 1, 2,\dots $}{
         Obtain expert advice  $ \xi_1(t),\ldots, \xi_N(t)$ for request $\sigma(t)$\;

         Select action $j \in \{1,\ldots, K\}$ with prob, $p_j(t) = (1-\eta) \sum_{i=1}^N \frac{w_i(t) \times \xi_i^j(t)}{W_t}  + \frac{\eta}{K}$\;
         
         \For{$i = 1,\dots,N$}{
          \begin{align*}
            \hat{x}_{i_t}(t) &=
            \begin{cases}
              \frac{x_{i_t}(t)}{d} , & \text{if}\ i_{t'}= j \text{ and } \sigma(t') \text{ is in history in position } d \\
              0, & \text{otherwise}
            \end{cases} \\
          w_i(t+1) & = w_i(t) \text{ exp } (\frac{-\eta\hat{\mathbf{x}}(t) \cdot \xi_i(t)}{K})\;
          \end{align*}
         }
      }
   } 
\caption{The \alec\ algorithm}
\label{alg:LeCaR}
\end{algorithm}

Next, we formalize the regret bound for the \alec\ algorithm. 
From the theoretical analysis of EXP4-DFDC, we have an optimal learning rate to choose for \alec\ with theoretical guarantees. 
Even though the learning rate suggested by Eq. (\ref{eq:opteta}) decreases with time, the decreasing learning rate works when the time horizon $T$ is known.
Furthermore, a decreasing learning rate is feasible when the environment is stationary and does not change over time once the distribution is learnt. But in dynamic settings such as for the cache page replacement problem, the system is not stationary over time. For this dynamic settings, decreasing learning rate will not work when the system is stationary for a period of time, but learning needs to be restarted as the environment change is detected. An an initial attempt, we set the learning rate at optimal value replacing $T$ with 1 and incorporating it in \alec. With this theoretically optimal learning rate, which depends on the cache size of the system and number of experts, we are able to get rid of the fixed choice of learning rate in \lec, and propose the new version called \alec.

\begin{corollary}
    Assume that \alec\ (Algorithm \ref{alg:LeCaR}) is run with a learning rate of $\eta =  \min(1, \sqrt{\frac{ K\ln N}{2T}})$, then the  expected regret of the algorithm can be upper bounded by,
    
\begin{equation}
R_{\text{\alec}}(T) \le 2\eta T + \frac{K\ln N}{\eta}.
\end{equation}
\label{cor:exp4dfdc}
\end{corollary}
 

\section{Conclusions}
\label{sec:conc}
In this paper, we have formulated a new MAB variant with delayed feedback and decaying cost (MAB-DFDC) applicable to the cache replacement problem. This version is different from all existing Bandit frameworks. We assume that feedback is always delayed, the cost decreases with increasing delay in feedback, and the regret vanishes over time if the expected delay crosses a certain threshold. 

As a solution, we propose the EXP4-DFDC algorithm and prove that expected regret is upper bounded by O(2$\sqrt{KT \ln N}$) for any learning rate $\eta$, where $K$ is the number of possible actions, $T$ - number of games, and $N$ is the number of experts. 

Finally, we show that the machine learning-based cache replacement algorithm \lec\  can be viewed as a simplified version of EXP4-DFDC. If the static learning rate of $\eta=0.45$ is replaced with the derived theoretical optimal learning rate, $\eta_{OPT} = \min(1, \sqrt{\frac{K\ln N}{2}})$, the associated regret will be upper bounded by $R_{\alec}(T)  \le 2 \sqrt{2KT\ln N }$. We refer this new version of \lec \ as \alec.

In this paper, we set the stage for theoretical analysis of reinforcement learning applied to the cache replacement problem. The multi-armed bandit analogy will help researchers find optimal hyperparameters for regret minimization and choose an appropriate model. Even small improvements in cache optimization may lead to a significant boost in storage systems performance.


\acks{This work was supported in part by NSF award CSR-1563883 and CNS-1956229. We acknowledge support for this project from Dr. Camilo Valdes and the rest of the \alec\ and \textsc{Cacheus} groups for their insightful feedback.}

\vskip 0.2in
\bibliography{main}

\end{document}